\documentclass[10pt]{article}

\usepackage{style/basic}

\definecolor{family}{HTML}{ECECEC}
\definecolor{pajamablue}{HTML}{E8F0FE}

\hypersetup{
    colorlinks=true,
    linkcolor=purple,
    urlcolor=teal,
    linktoc=all,
    citecolor=teal
}

\newtheorem*{theorem*}{Theorem}

\def \b0 {{\bf 0}}

\makeatletter

\newcommand{\changeoperator}[1]{%
  \csletcs{#1@saved}{#1@}%
  \csdef{#1@}{\changed@operator{#1}}%
}

\newcommand{\changed@operator}[1]{%
  \mathop{%
    \mathchoice{\textstyle\csuse{#1@saved}}
               {\csuse{#1@saved}}
               {\csuse{#1@saved}}
               {\csuse{#1@saved}}%
  }%
}

\makeatother

\newcommand\blfootnote[1]{%
  \begingroup
  \renewcommand\thefootnote{}\footnote{#1}%
  \addtocounter{footnote}{-1}%
  \endgroup
}

\newcommand{\EqualContribution}{\textsuperscript{*}Equal Contribution. Corresponding Authors: <thuang273, agoyal33>@wisc.edu}

\let\oldnl\nl
\newcommand{\nonl}{\renewcommand{\nl}{\let\nl\oldnl}}
\newif\ifsinglecolumn
\singlecolumntrue

\title{WARP: Weight-Space Analysis for Recovering Training Data Portfolios}

\author[$*\dagger$]{Tzu-Heng~Huang}
\author[$*\dagger$]{Aditya~Goyal}
\author[$\dagger$]{John~Cooper}
\author[$\dagger$]{Frederic~Sala}

\affil[$\dagger$]{University of Wisconsin-Madison}

\begin{document}
	\maketitle
	\blfootnote{\EqualContribution}

\begin{abstract}
Foundation models are routinely released to the public, yet the data recipes used to train them---such as domain mixture weights that determine how different sources are sampled---are rarely disclosed.
This creates an access asymmetry: researchers study the resulting models but lack visibility into the training distribution that produces them.
Prior works for inferring training data, such as membership inference, detect at the level of individual samples and thus cannot characterize the global composition of the training corpus.
We introduce WARP, a framework that recovers a fine-tuned model's training mixtures directly from its released weights.
WARP interpolates between the base and fine-tuned models using model merging, generating \emph{pseudo-checkpoints} that approximate the missing training trajectory and expose a geometric footprint of the training data in the weight space.
From these simulated footprints, WARP extracts geometric features and maps them to domain proportions using either a parameter-free softmax readout or an MLP projector trained on synthetic mixtures.
In controlled experiments with BERT and GPT-2, WARP recovers domain mixtures with an average MAE as low as $0.046$ and $0.104$ respectively, outperforming membership inference and a variant with access to the true training trajectory.
\end{abstract}
\footnotetext[1]{
This work appears in the ICML 2026 Workshop on Weight-Space Symmetries (WSS): from Foundations to Practical Applications.
}
\footnotetext[2]{
Our source code is available \href{https://github.com/SprocketLab/WARP}{here}.
}
\section{Introduction}
\label{sec:introduction}

%
Foundation models are trained on massive, heterogeneous corpora spanning web text, source code, and scientific literature.
One primary determinant of a model's ultimate capability is its \textbf{\emph{domain mixture}}: the proportions in which samples are drawn from these diverse sources~\cite{chen2026data}.
Recent studies have shown that tuning this data mixture is not merely a matter of ensuring knowledge variety, but of training efficiency---a well-designed mixture can yield models that substantially outperform naive sampling while requiring fewer optimization steps~\citep{xie2023doremi, ge2025r, chen2024aioli, fan2023doge}.

%
Yet the mixing recipes behind most frontier models remain proprietary and opaque.
This secrecy reflects a broader \textbf{\emph{access asymmetry}} in modern AI: general-purpose models are increasingly released to the public, while the carefully curated datasets used to produce them are not.
Such opacity creates structural barriers for the research community.
Practitioners who continually fine-tune a released model without knowing its original distribution risk unintended capability drift, and more broadly, recovering a model's data portfolio is useful for auditing data contamination~\citep{DBLP:journals/corr/abs-2308-08493, DBLP:journals/corr/abs-2311-06233, magar2022data} and for interpreting divergent model behaviors~\citep{singh2024rethinking}.

%
Breaking through this barrier, however, is far from straightforward.
Existing approaches to domain mixture optimization operate primarily in the \textbf{\emph{forward}} direction---\textbf{\emph{from data to model}}.
They either establish an optimized portfolio for a given dataset via gradient information~\citep{ge2025r, fan2023doge}, or train multiple proxy models that capture the relationship from mixtures to downstream performance~\citep{wen2026mixatlas, liu2025regmix, kang2024autoscale, shukor2026scaling}.
These natural designs make it hard to characterize already-released weights.
The alternative---working \textbf{\emph{backward from a model to its training data}}---has traditionally been framed as Membership Inference (MI)~\citep{fu2024membership, duan2024membership, mattern2023membership, shokri2017membershipinferenceattacksmachine}, which asks whether a particular sample appeared in the training pool, typically by comparing its attribution (e.g., loss or output logits) against a reference distribution.
However, MI is fundamentally a sample-level detection task and offers no holistic view of underlying domain proportions.

\begin{figure*}[t]
    \includegraphics[width=\linewidth]{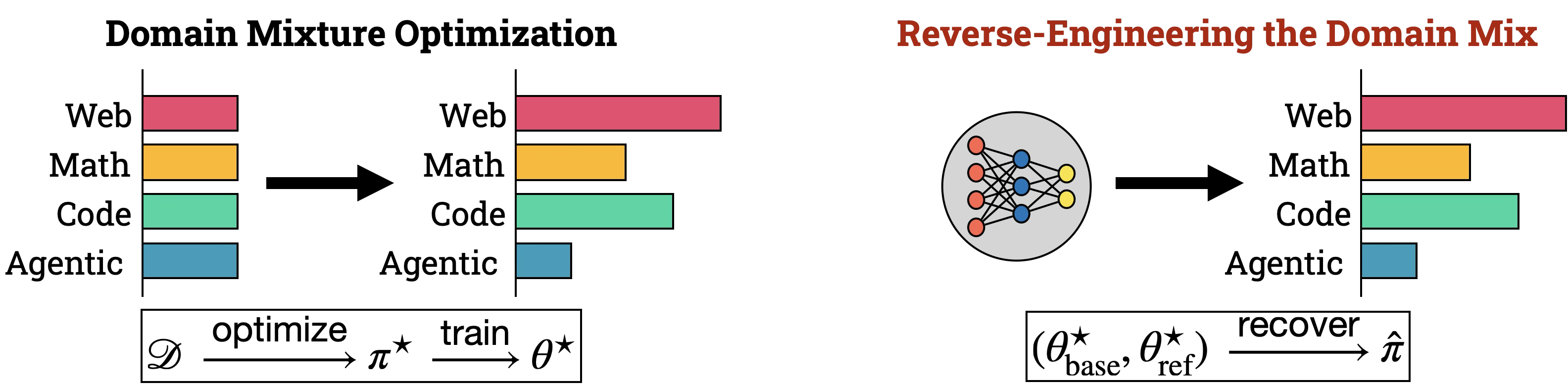}
    \caption{
    \textbf{Two directions of the domain mixture problem.}
    \textbf{Left:} the well-studied \emph{forward} problem---given a corpus, search for an optimized mixture $\pi^\star$ over domains (e.g., web, math, code, agentic data) and train a model $\theta^\star$ under it, whose final weights are then publicly released.
    \textbf{Right:} the \emph{inverse} problem we study---given only the released endpoints $(\theta^\star_{\mathrm{base}}, \theta^\star_{\mathrm{ref}})$ and a small probing dataset, with the corpus and true training trajectory withheld, recover an estimate $\hat{\pi}$ of the domain proportions that produced the fine-tuned model.
    }
    \label{fig:problem_setup}
\end{figure*}

%
One promising direction has recently appeared: using \emph{weight-space geometry}, the one artifact always available, to recover insights about training data~\cite{morris2025approximating, cazenavette2022dataset, huang2025evaluating}.
For example, the Mimic Score~\citep{huang2025evaluating} combines a directional vector in weight-space with training gradients to estimate sample utility for data curation.
These methods share a simple premise: \textbf{\emph{a model's parameters encode traces of its training history, and therefore of the data that produced it}}.
Yet they typically require access to \emph{intermediate} training checkpoints to reconstruct the learning trajectory---a luxury rarely available in practice.
For the vast majority of released models, only the final weights (e.g., \textsc{Instruct, Chat} versions) and the base model (e.g., pretrained-only) can be obtained: the origin and the destination, with no record of the path between them.

%
In this work, we break this asymmetry with \textbf{WARP} (\textbf{W}eight-space \textbf{A}nalysis for \textbf{R}ecovering Training Data \textbf{P}ortfolios), a framework that recovers a fine-tuned model's domain mixtures directly from its released weights.
We argue that \textbf{\emph{though the true training trajectory is lost, alternative interpolants between the two endpoints still expose usable structure in the surrounding weight-space geometry.}}
Specifically, we use model merging techniques to interpolate between the base and fine-tuned models, generating \textbf{\emph{pseudo-checkpoints}} that stand in for the unobserved parameter evolution.
From these simulated trajectories, their extracted geometric features allow us to capture a distributional footprint of the training data.
Given a small probing dataset, \textbf{(i)} a parameter-free softmax readout can recover a coarse mixture estimate; or \textbf{(ii)} a stronger variant uses synthetic mixtures to build an MLP for learning the projection from geometric features to predicted mixture.

We empirically validate WARP in a controlled setting by training BERT~\citep{devlin2019bert} and GPT-2~\citep{radford2019language} under known mixtures and recovering the proportions from each fine-tuned checkpoint.
Across forty experimental trials, WARP achieves a mean absolute error (MAE) as low as $0.046$ on BERT and $0.104$ on GPT-2 averaged across four text datasets, outperforming sample-level MI baselines and a variant that allows using the true training trajectory.
Moreover, the WARP estimator remains accurate on early-stop, converged, and overtrained checkpoints---mirroring today's post-training regime, where released models are routinely pushed beyond compute-optimal budgets, demonstrating its robustness to different training recipes.

We summarize our contributions as follows:
\begin{itemize}[nosep,topsep=0pt,leftmargin=*]
    \item \textbf{Closing the Access Asymmetry.}
    We formalize the recovery of a fine-tuned model's domain mixture from its weight-space footprints, relying only on artifacts available for most of the released models.
    \item \textbf{Simulated Training Trajectories.}
    We propose WARP, which sidesteps the need for intermediate checkpoints by reconstructing the missing path through model merging, then projects the resulting weight-space geometry to domain mixture estimates.
    \item \textbf{Strong Empirical Performance.}
    Across controlled fine-tuning runs, WARP outperforms sample-level MI baselines and even a variant that has access to the true training trajectory.
    \item \textbf{Robustness Across Training Recipes.}
    WARP remains accurate when recovering converged and overtrained checkpoints, demonstrating robustness to diverse training recipes.
\end{itemize}
\section{Related Work}
\label{app:related_work}

Our work sits at the intersection of three research threads: \textbf{(i)} domain mixture optimization, \textbf{(ii)} training data inference, and \textbf{(iii)} model merging techniques.

\paragraph{Domain Mixture Optimization.}
Establishing an ideal data mixture plays a critical role in training foundation models~\cite{chen2026data}.
Rather than relying on grid search to discover the best configuration, a recent line of work has developed principled methods for optimizing the data mixture~\cite{xie2023doremi, chen2024aioli, chen2026olmix}.
These methods broadly fall into two groups: (i) static mixing and (ii) dynamic reweighting.
Static methods fix the mixture before training, typically using proxy models to predict how mixture weights affect downstream performance---for example, by fitting a regression model that maps weights to performance~\cite{wen2026mixatlas, liu2025regmix, kang2024autoscale, shukor2026scaling}, or by leveraging gradient information from a held-out set to optimize the weights~\cite{fan2023doge}.
Dynamic methods instead adapt the mixture weights throughout training, for example by constructing a gradient-alignment matrix (R\&B)~\cite{ge2025r} or by casting data mixing as a multi-armed bandit problem~\cite{albalak2023efficient}.
\textbf{\emph{All of these methods operate in the forward direction: given the data, they find the best mixture.}}
In contrast, we address the reverse problem: given a model, we aim to recover the mixture that produced it.

\paragraph{Training Data Inference.}
Inferring properties of a model's training data, especially for large language models, is of growing interest, motivated both by privacy concerns---over the unauthorized use of copyrighted and personal information~\cite{DBLP:journals/corr/abs-2308-08493, mainidi2024}---and by the practical utility of extracting high-quality data insights for next-round data curation~\cite{huang2025evaluating}.
A common formulation is \emph{membership inference}~\cite{fu2024membership, duan2024membership, mattern2023membership, shokri2017membershipinferenceattacksmachine}, which detects whether a particular sample was used in training by comparing the model's behavior against a reference distribution, for instance, by examining the difference in output logits between an untuned and a fine-tuned model.
However, membership inference operates at the level of individual samples and therefore lacks a holistic view of how domains interact.
Even if the detected samples are aggregated to recover domain mixtures, the resulting estimate is confined to the probing dataset and remains suboptimal.
In this work, we propose a framework that \textbf{\emph{recovers a model's training trajectory in weight-space and leverages domain-level geometric features to recover the training mixture.}}

\paragraph{Model Merging Techniques.}
Model merging combines two or more models into a single fused model that inherits the capabilities of each~\cite{yang2026model, goddard-etal-2024-arcees, ilharco2022editing}.
It operates in weight space, where model weights can be permuted~\cite{ainsworth2022git, jordan2022repair}, interpolated~\cite{yadav2023ties, yu2024language}, stitched~\cite{he2024localize}, or averaged~\cite{wortsman2022model} to yield versatile skills.
Most work focuses on resolving conflicts between models to maximize the fused model's performance.
WARP departs from this view: rather than treating merging as a way to combine capabilities, \textbf{\emph{we use merging as a mechanism for simulating an unobserved training trajectory---constructing pseudo-checkpoints that stand in for the missing path between a base model and its fine-tuned descendant}}.
\section{Framework}
\label{sec:framework}

We propose \textbf{WARP} to recover the domain mixture of a fine-tuned model under two practical constraints: (i) the final weights and the corresponding base model are available, and (ii) the practitioner has access to a data source $S$ from which the fine-tuning data is drawn (e.g., the web datasets), and can freely sample a small set of labeled examples from it.
We argue that \textbf{\emph{the geometry traced between a base model and its fine-tuned descendant encodes a distributional footprint of the training data, and this footprint can be read out by measuring how samples from each domain align with the geometry}}.

\begin{figure*}[t]
    \includegraphics[width=\linewidth]{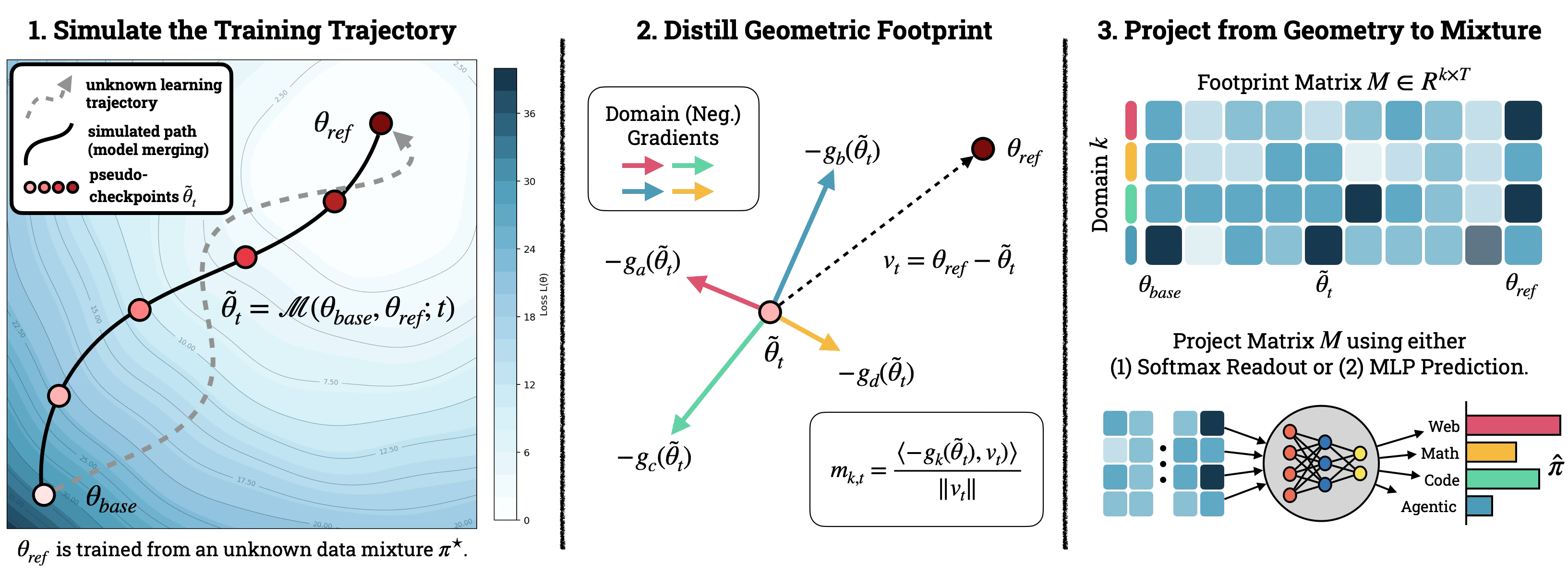}
    \caption{\textbf{Overview of WARP.} 
    \textbf{(1) Simulate the Training Trajectory:} since the true fine-tuning path is unobserved, we first approximate it with a sequence of pseudo-checkpoints obtained via model merging.
    \textbf{(2) Distill Geometric Footprint:} at each pseudo-checkpoint, we compute the Mimic Score by projecting the gradient of probing samples from each domain onto the direction pointing to reference model.
    This measures how strongly each domain aligns with the path actually taken during fine-tuning.
    \textbf{(3) Project from Geometry to Mixture:} the resulting footprint matrix is mapped to a predicted domain mixture, either via an unsupervised softmax readout or a supervised MLP projector trained on synthetic mixture-footprint pairs.
    }
    \label{fig:framework}
\end{figure*}

We begin with notation, then describe how we bypass the unobserved learning trajectory through simulation (\S\ref{sec:trajectory}), how the resulting weight-space geometry is converted into a domain-level signal (\S\ref{sec:footprint}), and finally how this signal is mapped to a predicted domain mixture (\S\ref{sec:mapping}).
We demonstrate our framework workflow in Figure~\ref{fig:framework}.
Moreover, we provide a unified algorithm in Appendix~\ref{app:algorithm}.

\paragraph{Notation.}
Let $\theta_{\text{base}}$ denote the parameters of a publicly available base model (pretrained-only), and $\theta_{\text{ref}}$ those of a released fine-tuned model derived from it.
Its fine-tuning data $D_{\text{ref}}$ is sampled from a source $S$ that spans $K$ domains, with unknown mixture proportions $\pi^\star = (\pi^\star_1, \dots, \pi^\star_K) \in \Delta^{K-1}$, where $\Delta^{K-1}$ is the probability simplex.
Our goal is to estimate $\pi^\star$ from $\theta_{\text{base}}$ and $\theta_{\text{ref}}$ alone, without access to $D_{\text{ref}}$ or any intermediate training checkpoint.
To reach this estimation, we assume the ability to sample examples from $S$ in our own way, allowing us to construct a small probing dataset $D_{\text{probe}} = \bigcup_{k=1}^{K} D_k$, where each $D_k = \{(x_i, y_i)\}_{i=1}^{n_k}$ contains samples with known domain label $k$ and $|D_{\text{probe}}| \ll |D_{\text{ref}}|$.
Crucially, $D_{\text{probe}}$ need not follow the same mixture as $D_{\text{ref}}$---it need only cover the $K$ domains of interest.
For any parameter vector $\theta$, we write each sample's gradient as $g_i(\theta) := \nabla_\theta \ell(x_i, y_i)$.

\subsection{Simulating the Training Trajectory}
\label{sec:trajectory}
To exploit the full geometric information along a training path, one would ideally have access to its intermediate checkpoints.
In practice, however, the true trajectory from $\theta_{\text{base}}$ to $\theta_{\text{ref}}$ is rarely released, leaving us with only its two endpoints.
To recover a usable approximation, we draw on recent advances in model merging~\citep{10.1145/3787849, goddard-etal-2024-arcees} and interpolate between these endpoints to construct a sequence of \textbf{\emph{pseudo-checkpoints}}: $\tilde{\theta}_t := \mathcal{M}\big(\theta_{\text{base}},\, \theta_{\text{ref}};\, t\big) \notag, $
where $\mathcal{M}$ is a merging operator.
For linear interpolation, TIES~\citep{yadav2023ties}, and other merging methods, it is common to have $\{\alpha_t\}_{t=1}^{T} \subset (0, 1)$ with
\begin{align*}
    \tilde{\theta}_t&= (1 - \alpha_t)\, \theta_{\text{base}} + \alpha_t\, \theta_{\text{ref}},
    \qquad t = 1, \dots, T,
\end{align*}
as a schedule of mixing coefficients that places each pseudo-checkpoint between the two endpoints.
Intuitively, $\alpha_t$ plays the role of a normalized training progress: values near $0$ yield an \emph{early-stage} state close to $\theta_{\text{base}}$, where the model has only begun to absorb the fine-tuning data, while values near $1$ yield a \emph{late-stage} state near $\theta_{\text{ref}}$, where most of the adaptation has already taken place.
Each $\tilde{\theta}_t$ thus serves as a proxy for the parameter state at an intermediate point along the simulated trajectory.

\subsection{Distilling Geometric Footprints}
\label{sec:footprint}
With a simulated trajectory in hand, we now extract a domain-level signal from pseudo-checkpoints' weight-space geometry.

\paragraph{Sample Alignment in Weight-space.}
The Mimic Score~\citep{huang2025evaluating} measures how well a sample's negative gradient aligns with a direction in the weight space pointing toward a more desirable parameter set: \emph{samples whose updates can push the model along this direction receive higher scores and are treated as higher-value training examples.}
We adapt this quality metric to our setting by treating each pseudo-checkpoint $\tilde{\theta}_t$ as the current state and $\theta_{\text{ref}}$ as the desired target, which together define a time-varying direction vector $v_t \;:=\; \theta_{\text{ref}} - \tilde{\theta}_t$.
For each probing sample $(x_i, y_i) \in D_{\text{probe}}$, the Mimic Score at step $t$ is the projection of its negative gradient onto the unit vector along $v_t$:
\begin{equation}
    m_{i, t} \;:=\; \frac{\langle -g_i(\tilde{\theta}_t),\, v_t \rangle}{\|v_t\|}.
\end{equation}
Intuitively, $m_{i, t}$ reflects how strongly sample $i$ would have pulled the model along the direction actually taken during fine-tuning.
\textbf{\emph{Samples from domains heavily represented in $D_{\text{ref}}$ should therefore yield relatively larger alignment than those from underrepresented ones}}.
Since $\|v_t\|$ shrinks as $\alpha_t \to 1$, raw scores at different steps are not directly comparable; we therefore apply min-max normalization to $\{m_{i, t}\}_i$ within each step $t$.

\paragraph{Domain-level Pooling.}
After normalization, we aggregate per-sample derived scores into a domain-level signal by averaging within each domain $k$ at each pseudo-checkpoint: 
\begin{align*}
    \bar{m}_{k, t} \;:=\; \frac{1}{|D_k|} \sum_{i \in D_k} m_{i, t}.
\end{align*}
Stacking these pooled scores across all $K$ domains and $T$ pseudo-checkpoints yields a feature matrix $M \in \mathbb{R}^{K \times T}$, which serves as the \textbf{\emph{geometric footprint}} of the simulated trajectory: \textbf{\emph{row $k$ traces how strongly domain $k$ aligns with the fine-tuning direction as the model evolves from $\theta_{\text{base}}$ toward $\theta_{\text{ref}}$}}.

\subsection{Mapping from Geometry to Mixture}
\label{sec:mapping}
The final step is to translate the geometric footprint $M$ into a predicted mixture $\hat{\pi} \in \Delta^{K-1}$.
We instantiate this mapping in two ways: using an \emph{unsupervised} variant that reads the mixture directly off $M$, or leveraging a \emph{supervised} variant that learns the mapping from synthetic labeled pairs.

\paragraph{Unsupervised Readout.}
The simplest mapping converts $M$ into a mixture estimate without any learnable parameters, by averaging the pooled scores across pseudo-checkpoints and applying a softmax over domains:
\begin{equation}
    \hat{\pi}_k \;=\; \frac{\exp(\dot{m}_k / \tau)}{\sum_{k'=1}^{K} \exp(\dot{m}_{k'} / \tau)},
    \quad \dot{m}_k \;=\; \frac{1}{T} \sum_{t=1}^{T} \bar{m}_{k, t},
\end{equation}
where $\tau > 0$ is a temperature controlling the sharpness of the predicted distribution.
This readout is \emph{fast and label-free}, but by collapsing the temporal dimension it cannot exploit how domain influence varies across early and late learning.

\paragraph{Supervised Projection.}
Our second variant trains a projector using synthetic pairs constructed from the same data source $S$.
Concretely, we draw $J$ random mixtures $\pi^{(j)}$ from a distribution $\Pi$ over the probability simplex, designed so that $\|\pi^{(j)} - \frac{1}{K}\mathbf{1}\|_1$ is uniform.
For each sampled mixture, we draw a new dataset from $S$ accordingly and fine-tune the base model on it to obtain a synthetic reference model, then compute its footprint $M^{(j)}$ using the probing dataset $D_{\text{probe}}$.
Moreover, since each fine-tune is only a means of \emph{generating training signal} for the projector---not a model intended for downstream use---\emph{we do not need to optimize these mixtures}.
\textbf{\emph{That is, cheap, short fine-tuning runs are acceptable, making synthetic pairs efficient to produce at scale}}.
On these pairs, we train an MLP with a softmax output layer, $f_\phi : \mathbb{R}^{K \times T} \to \Delta^{K-1}$, that maps the full feature matrix to a predicted mixture $\hat{\pi} = f_\phi(M)$, using a standard regression loss against the synthetic ground-truth mixture $\pi^{(j)}$.
By operating on the full $M$ rather than its temporal average, this learns to weight different learning stages adaptively, capturing how domain influence evolves along the simulated trajectory.

\section{Experiments}
\label{sec:experiments}


We evaluate WARP in a controlled setting where the ground-truth domain mixture is known, allowing us to directly measure the recovery effectiveness.
Our goal is to verify:
\begin{enumerate}[nosep,topsep=0pt,leftmargin=*]
    \item \textbf{Effectiveness of Weight-Space Geometry.}
    Recovering domain mixtures from weight-space geometry substantially outperforms random guessing and sample-level membership inference baselines.
    \item \textbf{Efficiency of Unsupervised Readout.}
    Even without any learned projector, a parameter-free softmax readout can recover domain proportions accurately.
    \item \textbf{Simulated Paths Suffice.}
    Pseudo-checkpoints obtained via model merging can simulate the true learning path, matching---and sometimes outperforming---the variant granted access to real checkpoints.
    \item \textbf{Robustness Across Training Recipes.}
    WARP remains accurate on both converged and overtrained checkpoints, mirroring the post-training regime where current models are pushed beyond compute-optimal budgets.
\end{enumerate}

\paragraph{Setups.}
We use four text datasets to evaluate WARP, spanning different domain structures: 
(i) \textsc{SNLI}~\citep{young-etal-2014-image} (3 classes, natural language inference), 
(ii) \textsc{AGNews}~\citep{zhang2015character} (4 classes, news topic classification), 
(iii) \textsc{Yelp}~\citep{zhang2015character} (5 classes, review sentiment), and 
(iv) \textsc{Yahoo}~\citep{zhang2015character} (10 classes, topic classification).
We treat each class as a representative domain.
For every dataset, we sample $40$ ground-truth mixtures $\pi^\star$ over $\Delta^{K-1}$ to construct a $5,000$-sized dataset and fine-tune both \textsc{BERT-base} and \textsc{GPT-2-Small} on data drawn according to each $\pi^\star$ to obtain $\theta_{\text{ref}}$, using the corresponding pretrained weights as $\theta_{\text{base}}$.
The probing dataset $D_{\text{probe}}$ is drawn separately from the same source $S$, and $|D_{\text{probe}}| = 2{,}500$ examples sampled uniformly per domain.
Pseudo-checkpoints are constructed at $T = 15$ evenly spaced interpolation steps between $\theta_{\text{base}}$ and $\theta_{\text{ref}}$.
Recovery performance is reported using mean absolute error (MAE) between $\hat{\pi}$ and $\pi^\star$, averaged across the $40$ trials.
We provide our training configurations in Appendix~\ref{app:experimental_details}.

\begin{table*}[t!]
\centering
\caption{Domain mixture recovery error (MAE between predicted $\hat{\pi}$ and ground-truth $\pi^\star$, averaged over $40$ trials) on BERT and GPT-2-Small across datasets. Lower is better. Best results within each model block are \textbf{bolded}; second-best are \underline{underlined}.}
\label{tab:combined_mae}
\definecolor{lightblue}{RGB}{220,235,250}

\begin{tabular}{@{} l
*{4}{S[table-format=1.3, detect-weight]}
>{\columncolor{lightblue}}S[table-format=1.3, detect-weight]
*{4}{S[table-format=1.3, detect-weight]}
>{\columncolor{lightblue}}S[table-format=1.3, detect-weight]
@{}}

\toprule
& \multicolumn{5}{c}{\textbf{BERT}} 
& \multicolumn{5}{c}{\textbf{GPT-2-Small}} \\
\cmidrule(lr){2-6} \cmidrule(lr){7-11}

\textbf{Method}
& {\textbf{SNLI}} & {\textbf{AGNews}} & {\textbf{Yelp}} & {\textbf{Yahoo}} & {\textbf{Avg.}}
& {\textbf{SNLI}} & {\textbf{AGNews}} & {\textbf{Yelp}} & {\textbf{Yahoo}} & {\textbf{Avg.}} \\

& {\scriptsize (3 cls.)} & {\scriptsize (4 cls.)} & {\scriptsize (5 cls.)} & {\scriptsize (10 cls.)} & {\textbf{MAE}}
& {\scriptsize (3 cls.)} & {\scriptsize (4 cls.)} & {\scriptsize (5 cls.)} & {\scriptsize (10 cls.)} & {\textbf{MAE}} \\

\midrule

\rowcolor[gray]{0.9}
\multicolumn{11}{l}{\textbf{\textit{Baselines}}} \\
Random Guess     & 0.294 & 0.239 & 0.192 & {0.090} & 0.204 & 0.294 & 0.239 & 0.192 & {0.090} & 0.204 \\
Centroid Guess   & 0.220 & 0.179 & 0.149 & {0.069} & 0.154 & 0.220 & 0.179 & 0.149 & {0.069} & 0.154 \\
Sample-level MI  & {\bfseries 0.074} & 0.084 & 0.064 & {\bfseries 0.029} & 0.063 & 0.214 & 0.200 & {\bfseries 0.091} & {\bfseries 0.059} & 0.141 \\

\rowcolor[gray]{0.9}
\multicolumn{11}{l}{\textbf{\textit{Our Method: Unsupervised Softmax}}} \\
\quad Real Trajectory & 0.295 & 0.321 & 0.180 & {0.061} & 0.214 & 0.205 & 0.152 & 0.168 & {0.092} & 0.154 \\
\quad SLERP           & 0.202 & 0.142 & 0.106 & {0.068} & 0.130 & 0.362 & 0.308 & 0.262 & {0.125} & 0.264 \\
\quad LERP            & 0.091 & 0.118 & 0.086 & {0.049} & 0.086 & 0.137 & 0.185 & 0.131 & {0.064} & 0.130 \\
\quad TIES            & 0.091 & 0.118 & 0.087 & {0.049} & 0.086 & 0.138 & 0.175 & 0.130 & {0.064} & 0.127 \\

\rowcolor[gray]{0.9}
\multicolumn{11}{l}{\textbf{\textit{Our Method: Supervised 2-Layer MLP}}} \\
\quad Real Trajectory & 0.142 & 0.049 & 0.048 & {0.055} & 0.074 & 0.151 & {\underline{0.127}} & 0.137 & {0.069} & 0.121 \\
\quad SLERP           & 0.209 & 0.110 & 0.094 & {0.088} & 0.130 & 0.232 & 0.211 & 0.173 & {0.075} & 0.172 \\
\quad LERP            & 0.092 & {\bfseries 0.031} & {\bfseries 0.022} & 0.042 & {\underline{0.047}} & {\bfseries 0.106} & {\bfseries 0.115} & {\underline{0.129}} & {0.064} & {\bfseries 0.104} \\
\quad TIES            & {\underline{0.087}} & {\underline{0.034}} & {\underline{0.023}} & {\underline{0.040}} & {\bfseries 0.046} & {\underline{0.108}} & 0.130 & 0.135 & {\underline{0.063}} & {\underline{0.109}} \\

\bottomrule
\end{tabular}%
\end{table*}

\paragraph{Baselines.}
We compare WARP against \emph{three} baselines:
\begin{itemize}[nosep,topsep=0pt,leftmargin=*]
    \item \textbf{Random Guess}: A mixture sampled from $\Delta^{K-1}$, serving as an uninformed lower bound.
    \item \textbf{Centroid Guess}: A uniform mixture $\hat{\pi} = \frac{1}{K}\mathbf{1}$, acting as the best constant predictor without observing $\theta_{\text{ref}}$.
    \item \textbf{Sample-level MI}: 
    We score each probing example by its loss reduction from $\theta_{\text{base}}$ to $\theta_{\text{ref}}$, sweep the reduction threshold to \emph{maximize detection accuracy}, and aggregate the resulting per-domain membership rates into a mixture estimate.
    We treat this as the closest sample-based dataset recovery to our weight-space approach.
\end{itemize}

To isolate the effect of \emph{simulation}, we further include an \emph{oracle variant} of WARP that operates on real intermediate checkpoints saved during fine-tuning, in place of pseudo-checkpoints from merging.
For the merging operator $\mathcal{M}$, we evaluate three choices: \textbf{SLERP} (spherical interpolation), \textbf{LERP} (linear interpolation), and \textbf{TIES}~\citep{yadav2023ties}.
\paragraph{Main Results.}
Table~\ref{tab:combined_mae} reports WARP's recovery accuracy, supporting three observations.
\textbf{First, weight-space geometry carries a strong signal for estimation.}
WARP's best supervised variant attains $0.046$ MAE on BERT and $0.104$ on GPT-2-Small, reducing the strongest baseline's error by $35\%$ and $30\%$ respectively---well below random guessing and sample-level membership inference.
\textbf{Second, unsupervised readout is competitive.}
Even without a learned projector---which would require collecting an additional synthetic training set---the parameter-free softmax readout already outperforms all baselines and closes most of the gap to the supervised variant.
\textbf{Third, simulated trajectories match or surpass real ones.}
On BERT, supervised LERP and TIES both reach $0.048$ avg.\ MAE versus $0.080$ for the oracle on real checkpoints; on GPT-2-Small, LERP reaches $0.117$ versus $0.138$.
We attribute this to a \emph{smoothness gap}: real fine-tuning paths can be noisy---stochastic mini-batches, learning-rate warmup, and curriculum effects inject variance into per-step alignment scores.
LERP and TIES instead create a clean monotone path from $\theta_{\text{base}}$ to $\theta_{\text{ref}}$, yielding a smoother footprint matrix $M$ from which the 2-layer MLP can more easily extract the mixture.
Overall, we demonstrate the feasibility of leveraging weight-space geometry and the dataset footprints to reverse-engineer the training mixture.
\begin{table*}[t!]
\centering
\caption{
Domain mixture recovery error on GPT-2-Small (using AG News and TIES merging) across different training stages.
Both methods remain robust at recovering different types of checkpoint, with the supervised MLP consistently outperforming all baselines.
Lower is better.
}
\label{tab:mse_mae_checkpoints}
\small
\begin{tabular}{@{} l *{6}{S[table-format=1.3, detect-weight]} @{}}
\toprule
& \multicolumn{2}{c}{\textbf{Short Runs} \scriptsize(9 ep.)} & \multicolumn{2}{c}{\textbf{Converged} \scriptsize(12 ep.)} & \multicolumn{2}{c}{\textbf{Overtrained} \scriptsize(18 ep.)} \\
\cmidrule(lr){2-3} \cmidrule(lr){4-5} \cmidrule(lr){6-7}
\textbf{Method} & {\textbf{MSE}} & {\textbf{MAE}} & {\textbf{MSE}} & {\textbf{MAE}} & {\textbf{MSE}} & {\textbf{MAE}} \\
\midrule
\rowcolor[gray]{0.9} \multicolumn{7}{l}{\textbf{\textit{Baselines}}} \\
Random Guess         & 0.095 & 0.239 & 0.095 & 0.239 & 0.095 & 0.239 \\
Centroid Guess       & 0.048 & 0.179 & 0.048 & 0.179 & 0.048 & 0.179 \\
Sample-level MI      & 0.068 & 0.200 & 0.068 & 0.200 & 0.068 & 0.200 \\
\midrule
\rowcolor[gray]{0.9} \multicolumn{7}{l}{\textbf{\textit{Our Methods}}} \\
\quad Unsupervised Softmax     & 0.053 & 0.175 & 0.060 & 0.183 & 0.064 & 0.193 \\
\quad Supervised 2-Layer MLP   & \bfseries 0.033 & \bfseries 0.130 & \bfseries 0.040 & \bfseries 0.147 & \bfseries 0.030 & \bfseries 0.124 \\
\bottomrule
\end{tabular}
\end{table*}

\paragraph{Different Training Recipes.}
An interesting question when deploying our method is which reference checkpoint to distill alignment matrices from, since the right choice is often unclear in practice---practitioners may stop training early to save compute, train to convergence, or inadvertently overtrain.
Table~\ref{tab:mse_mae_checkpoints} evaluates this sensitivity by extracting geometric footprints at three different checkpoints along the GPT-2-Small fine-tuning trajectory on AG News: including short runs (9 epochs, mirroring early stopping), converged checkpoints (12 epochs), and overtrained checkpoints (18 epochs, well past convergence).
Across all three stages, both our unsupervised softmax and supervised MLP variants outperform the baselines, with the supervised MLP achieving the lowest recovery error.
This demonstrates our framework's robustness when facing diverse training recipes.
\section{Conclusion}
\label{sec:conclusion}

In this work, we introduce WARP, a framework that recovers a fine-tuned model's training domain mixture from its weights alone.
By using model merging to simulate the missing training trajectory, WARP extracts a geometric footprint of the training data and maps it to domain proportions.
In controlled experiments, WARP achieves MAE as low as 0.046 on BERT and 0.104 on GPT-2, outperforming membership inference baselines and even an oracle with access to the true trajectory.
Overall, we demonstrate the feasibility of recovering a model's training distribution from its weight-space geometry---a step toward greater transparency in an era where model weights are shared but data recipes are not.

\bibliographystyle{achemso}
\bibliography{reference}

\newpage
\appendix

\newpage
\appendix
\onecolumn

\section*{Appendix Roadmap}
Our appendix is structured as follows.
We use Appendix~\ref{app:algorithm} to summarize our full framework in pseudocode.
Appendix~\ref{app:experimental_details} documents the experimental setup and training configurations.
Finally, Appendix~\ref{app:future_work} outlines directions for future work.

\section{Algorithm}
\label{app:algorithm}

Next, we summarize the full WARP procedure in Algorithm~\ref{alg:warp}.
WARP proceeds in three stages: simulating a trajectory between $\theta_{\text{base}}$ and $\theta_{\text{ref}}$ via model merging (Sec.~\ref{sec:trajectory}), distilling each pseudo-checkpoint into a domain-level geometric footprint via Mimic Scores (Sec.~\ref{sec:footprint}), and mapping the resulting footprint matrix to a predicted mixture through either an unsupervised softmax readout or a supervised projector trained on synthetic pairs (Sec.~\ref{sec:mapping}).

\begin{algorithm}[h!]
\caption{\textbf{WARP}: Weight-Space Analysis for Recovering Data Portfolios}
\label{alg:warp}
\begin{algorithmic}[1]
\Require 
    Base model $\boldsymbol{\theta}_{\text{base}}$, 
    fine-tuned model $\boldsymbol{\theta}_{\text{ref}}$, 
    data source $S$ over $K$ domains, 
    merging operator $\mathcal{M}$ with schedule $\{\alpha_t\}_{t=1}^{T}$, 
    mode $\in \{\textsc{Unsup}, \textsc{Sup}\}$, 
    temperature $\tau$, 
    (\textsc{Sup} only) mixture distribution $\Pi$ and \# pairs $J$.
\Ensure 
    Estimated domain mixture $\hat{\boldsymbol{\pi}} \in \Delta^{K-1}$.
\State \textbf{Initialize:} probing dataset $D_{\text{probe}} = \bigcup_{k=1}^{K} D_k$ sampled from $S$ with known labels
\Statex
\Statex \textit{// Stage 1: Simulate trajectory \quad // Stage 2: Distill footprint}
\For{$t = 1$ \textbf{to} $T$}
    \State $\tilde{\boldsymbol{\theta}}_t \gets \mathcal{M}(\boldsymbol{\theta}_{\text{base}},\, \boldsymbol{\theta}_{\text{ref}};\, \alpha_t)$, \quad $\mathbf{v}_t \gets \boldsymbol{\theta}_{\text{ref}} - \tilde{\boldsymbol{\theta}}_t$ \hfill \Comment{Pseudo-checkpoint and direction}
    \State $m_{i,t} \gets \langle -\nabla \ell(x_i, y_i;\, \tilde{\boldsymbol{\theta}}_t),\, \mathbf{v}_t \rangle \,/\, \|\mathbf{v}_t\|$ \quad for $(x_i, y_i) \in D_{\text{probe}}$ \hfill \Comment{Mimic Score}
    \State Min-max normalize $\{m_{i,t}\}_i$; \quad $\bar{m}_{k,t} \gets \frac{1}{|D_k|} \sum_{i \in D_k} m_{i,t}$ \hfill \Comment{Pool by domain}
\EndFor
\State $M \gets [\bar{m}_{k,t}]_{K \times T}$ \hfill \Comment{Geometric footprint}
\Statex
\Statex \textit{// Stage 3: Map footprint to mixture}
\If{Mode is \textsc{Unsup}}
    \State $\hat{\pi}_k \gets \mathrm{softmax}_k\!\left(\tfrac{1}{T} \sum_{t} \bar{m}_{k,t} \,/\, \tau\right)$ \hfill \Comment{Label-free readout}
\Else
    \For{$j = 1$ \textbf{to} $J$}
        \State Sample $\boldsymbol{\pi}^{(j)} \sim \Pi$, draw $D^{(j)}$ from $S$ with mixture $\boldsymbol{\pi}^{(j)}$, fine-tune $\boldsymbol{\theta}_{\text{base}}$ on $D^{(j)}$ to get $\boldsymbol{\theta}^{(j)}_{\text{ref}}$
        \State $M^{(j)} \gets$ Stages~1--2 with $(\boldsymbol{\theta}_{\text{base}},\, \boldsymbol{\theta}^{(j)}_{\text{ref}},\, D_{\text{probe}})$
    \EndFor
    \State Train $f_\phi : \mathbb{R}^{K \times T} \to \Delta^{K-1}$ on $\{(M^{(j)},\, \boldsymbol{\pi}^{(j)})\}_{j=1}^{J}$; \quad $\hat{\boldsymbol{\pi}} \gets f_\phi(M)$
\EndIf
\State \Return $\hat{\boldsymbol{\pi}}$
\end{algorithmic}
\end{algorithm}
\section{Experimental Details}
\label{app:experimental_details}

Here we provide additional details of our experimental setup.
We discuss (i) how reference models are constructed, (ii) how the probing dataset is sampled, (iii) the supervised projector training pipeline, and (iv) the computational resources used.

\paragraph{Constructing Reference Models.}
We validate WARP in a controlled setting where the ground-truth mixture $\pi^\star$ is known.
For each dataset, we sample $40$ mixtures $\{\pi^\star_i\}_{i=1}^{40}$ from the simplex $\Delta^{K-1}$.
For each $\pi^\star_i$, we draw $5{,}000$ training examples whose per-domain proportions follow $\pi^\star_i$, and fine-tune both \textsc{BERT} and \textsc{GPT-2-Small} for $9$ epochs starting from their publicly released pretrained weights, which we use as $\theta_{\text{base}}$.
For each reference model, we sweep the learning rate over $\{6\mathrm{e}{-}6,\ 1\mathrm{e}{-}5,\ 3\mathrm{e}{-}5,\ 5\mathrm{e}{-}5,\ 1\mathrm{e}{-}4\}$ and select the value that minimizes training loss to obtain $\theta_{\text{ref}}$.
This yields $40$ reference models per (dataset, architecture) pair.

\paragraph{Creating Probing Dataset.}
WARP makes a simple assumption about the domain distribution of the probing set; it requires only that $D_{\text{probe}}$ shares the same $K$ domains as $D_{\text{ref}}$.
We sample $|D_{\text{probe}}| = 2{,}500$ examples from $S$, with $2{,}500/K$ examples drawn uniformly per domain.
The probing dataset is fixed once per (dataset, architecture) pair and reused across all $40$ mixtures.
For each probing example, we compute its alignment score from the per-sample gradient and the directional vector pointing from each pseudo-checkpoint toward $\theta_{\text{ref}}$.

\paragraph{Training Projector.}
For the supervised variant, the projector is a $2$-layer MLP with ReLU activation, and a softmax output layer that produces a vector on $\Delta^{K-1}$.
The input is the matrix of domain-level alignment statistics aggregated from $D_{\text{probe}}$, and the target is the ground-truth mixture $\pi^\star$.
We train it with a learning rate $1\mathrm{e}{-}4$ for $200$ epochs.
Since we have $40$ (mixture, reference-model) pairs per (dataset, architecture), we report results under $5$-fold cross-validation: in each fold, $32$ pairs train the projector and the held-out $8$ pairs are used for evaluation.
All the experiments are conducted on an NVIDIA RTX A6000 GPU.

\paragraph{Evaluation.}
We report MAE between $\hat{\pi}$ and $\pi^\star$, averaged over the $40$ ground-truth mixtures in Table~\ref{tab:combined_mae}.
We additionally report mean squared error (MSE) in Table~\ref{tab:combined_mse}.

\begin{table*}[t!]
\centering
\small
\caption{
Domain mixture recovery error (MSE between predicted $\hat{\pi}$ and ground-truth $\pi^\star$, averaged over $40$ trials) on BERT and GPT-2-Small across datasets.
Lower is better.
Best results within each model block are \textbf{bolded}; second-best are \underline{underlined}.
}
\label{tab:combined_mse}
\definecolor{lightblue}{RGB}{220,235,250}

\resizebox{\textwidth}{!}{%
\begin{tabular}{@{} l
*{4}{S[table-format=1.3, detect-weight]}
>{\columncolor{lightblue}}S[table-format=1.3, detect-weight]
*{4}{S[table-format=1.3, detect-weight]}
>{\columncolor{lightblue}}S[table-format=1.3, detect-weight]
@{}}
\toprule
& \multicolumn{5}{c}{\textbf{BERT}} & \multicolumn{5}{c}{\textbf{GPT-2-Small}} \\
\cmidrule(lr){2-6} \cmidrule(lr){7-11}

\textbf{Method}
& {\textbf{SNLI}} & {\textbf{AGNews}} & {\textbf{Yelp}} & {\textbf{Yahoo}} & {\textbf{Avg.}}
& {\textbf{SNLI}} & {\textbf{AGNews}} & {\textbf{Yelp}} & {\textbf{Yahoo}} & {\textbf{Avg.}} \\

& {\scriptsize (3 cls.)} & {\scriptsize (4 cls.)} & {\scriptsize (5 cls.)} & {\scriptsize (10 cls.)} & {\textbf{MSE}}
& {\scriptsize (3 cls.)} & {\scriptsize (4 cls.)} & {\scriptsize (5 cls.)} & {\scriptsize (10 cls.)} & {\textbf{MSE}} \\

\midrule

\rowcolor[gray]{0.9}
\multicolumn{11}{l}{\textbf{\textit{Baselines}}} \\
Random Guess     & 0.140 & 0.095 & 0.066 & {0.014} & 0.079 & 0.140 & 0.095 & 0.066 & 0.014 & 0.079 \\
Centroid Guess   & 0.072 & 0.048 & 0.032 & {0.007} & 0.040 & 0.072 & 0.048 & \underline{0.032} & \textbf{0.007 }& 0.040 \\
Sample-level MI       & {\bfseries 0.009} & 0.012 & 0.007 & \textbf{{0.001}} & 0.007 & 0.073 & 0.068 & {\bfseries 0.017} & \underline{0.008} & 0.042 \\

\rowcolor[gray]{0.9}
\multicolumn{11}{l}{\textbf{\textit{Our Method: Unsupervised Softmax}}} \\
\quad Real Trajectory & 0.156 & 0.163 & 0.062 & {0.007} & 0.097 & 0.070 & 0.046 & 0.058 & 0.019 & 0.048 \\
\quad SLERP           & 0.071 & 0.029 & 0.019 & {0.008} & 0.032 & 0.197 & 0.163 & 0.121 & 0.036 & 0.130 \\
\quad LERP            & {\underline{0.013}} & 0.021 & 0.013 & {0.004} & 0.013 & 0.036 & 0.062 & {\underline{0.032}} & \underline{0.008} & 0.035 \\
\quad TIES            & {\underline{0.013}} & 0.021 & 0.013 & {0.004} & 0.013 & 0.036 & 0.053 & {\underline{0.032}} & \underline{0.008} & 0.032 \\

\rowcolor[gray]{0.9}
\multicolumn{11}{l}{\textbf{\textit{Our Method: Supervised 2-Layer MLP}}} \\
\quad Real Trajectory & 0.040 & {\underline{0.005}} & {\underline{0.005}} & {0.006} & 0.014 & 0.040 & {\underline{0.030}} & 0.040 & 0.009 & 0.030 \\
\quad SLERP           & 0.070 & 0.018 & 0.018 & {0.014} & 0.034 & 0.080 & 0.077 & 0.053 & 0.011 & 0.055 \\
\quad LERP            & 0.017 & {\bfseries 0.002} & {\bfseries 0.001} & {0.004} & {\underline{0.006}} & {\bfseries 0.022} & {\bfseries 0.024} & 0.036 & \underline{0.008} & {\bfseries 0.023} \\
\quad TIES            & 0.014 & {\bfseries 0.002} & {\bfseries 0.001} & \underline{0.003} & {\bfseries 0.005} & {\underline{0.023}} & 0.033 & 0.037 & \underline{0.008} & \underline{0.025} \\

\bottomrule
\end{tabular}
}
\end{table*}

\section{Future Work}
\label{app:future_work}

WARP provides the \textbf{\emph{first}} mechanism for recovering the domain mixture of a fine-tuned model from its released weights alone.
We outline several directions that can extend its scope, including (i) interpreting the learned projector, (ii) strengthening the unsupervised readout, and (iii) scaling to large-scale LLMs.

\paragraph{Interpretability of the Learned Projector.}
The supervised variant of WARP encodes, in the weights of its MLP projector, an \emph{implicit} model of how Mimic Scores across pseudo-checkpoints map to domain proportions.
Understanding this mapping would clarify \emph{which} stages of the simulated trajectory carry the strongest signal for each domain, and \emph{why}.
A starting point is to replace the MLP with shallower, fully linear projectors, whose coefficients can be read directly as per-stage importance weights.
Combined with a larger set of synthetic pseudo-checkpoints and explicit regularization (e.g., sparsity or smoothness across $t$), this could begin to expose the temporal structure of domain influence as the model evolves from $\theta_{\text{base}}$ toward $\theta_{\text{ref}}$.

\paragraph{Advanced Unsupervised Readouts.}
The current unsupervised variant uses two assumptions: that uniform averaging of pooled Mimic Scores across pseudo-checkpoints is a \emph{reliable} aggregation, and that the resulting per-domain quantities behave like logits suitable for a softmax.
A more principled alternative would treat the readout as inference in a graphical model over pseudo-checkpoints, where each step contributes a noisy vote on the underlying mixture and \emph{weak supervision} techniques infer per-stage reliabilities without labels.
Such unsupervised techniques could adaptively down-weight uninformative stages, potentially narrowing the gap to the supervised variant while preserving the label-free setting.

\paragraph{Scaling to Large-Scale LLMs.}
Finally, our controlled experiments fine-tune BERT and GPT-2 under known mixtures, which isolates the recovery problem but leaves open how WARP behaves at the scale of modern LLM training.
LLMs are typically adapted through long, multi-stage pipelines---continued mid-training, supervised fine-tuning, and preference optimization---over corpora spanning dozens of domains.
Extending WARP to these regimes raises two questions in particular: whether linear interpolation remains a useful proxy for trajectories that traverse qualitatively different objectives, and how the geometric footprint behaves when domains are numerous and more complex.
These directions will help extend WARP from a controlled diagnostic tool toward a recovery tool applicable to frontier models.

\end{document}